\documentclass[letterpaper,journal]{IEEEtran}
\usepackage{amsmath,amsfonts}
\usepackage{algorithmic}
\usepackage{algorithm}
\usepackage{array}
\usepackage[caption=false,font=normalsize,labelfont=sf,textfont=sf]{subfig}
\usepackage{textcomp}
\usepackage{stfloats}
\usepackage{url}
\usepackage{verbatim}
\usepackage{graphicx}
\usepackage{cite}

\begin{document}

\title{Improving the Diversity of Bootstrapped DQN by Replacing Priors with Noise}


\author{Li Meng$^1$, Morten Goodwin$^{2,3}$, Anis Yazidi$^{3,4,5}$, Paal Engelstad$^1$\\
$^1$University of Oslo\\
$^2$Centre for Artificial Intelligence Research, University of Agder\\
$^3$Oslo Metropolitan University\\
$^4$Norwegian University of Science and Technology\\
$^5$Oslo University Hospital}


\IEEEpubid{DOI: 10.1109/TG.2022.3185330~\copyright~2022 IEEE}


\maketitle

\begin{abstract}
Q-learning is one of the most well-known Reinforcement Learning algorithms. There have been tremendous efforts to develop this algorithm using neural networks. Bootstrapped Deep Q-Learning Network is amongst them. It utilizes multiple neural network heads to introduce diversity into Q-learning. Diversity can sometimes be viewed as the amount of reasonable moves an agent can take at a given state, analogous to the definition of the exploration ratio in RL. Thus, the performance of Bootstrapped Deep Q-Learning Network is deeply connected with the level of diversity within the algorithm. In the original research, it was pointed out that a random prior could improve the performance of the model. In this article, we further explore the possibility of replacing priors with noise and sample the noise from a Gaussian distribution to introduce more diversity into this algorithm. We conduct our experiment on the Atari benchmark and compare our algorithm to both the original and other related algorithms. The results show that our modification of the Bootstrapped Deep Q-Learning algorithm achieves significantly higher evaluation scores across different types of Atari games. Thus, we conclude that replacing priors with noise can improve Bootstrapped Deep Q-Learning's performance by ensuring the integrity of diversities.
\end{abstract}


\section{Introduction}
Reinforcement Learning (RL) is a sub-field of Artificial Intelligence (AI) that studies intelligent behaviors through indicative reward signals. The term "intelligence" can be interpreted in different ways. There were efforts towards describing intelligence in mathematical formulas \cite{LeggUniversal}, yet there is no strictly definable quantitative measure about the level of intelligence that one agent possesses. Intelligence, given its multi-pronged nature, can involve a variety of skills that we consider as intelligent behaviors. Thus, focusing on one particular intelligent aspect of an agent can be a progressive approach to research AI \cite{schaul2011measuring}.

RL is suitable in studying games because it requires the input of reward signals, and games typically have well-defined reward mechanisms to guide a player towards goals. On the other hand, it is often challenging to build a simulator that gives adequate reward signals and precisely describes the complexity of the real world. Games also have finite action and time sequences, enabling the quantitative measure of the intelligence that would not be applicable otherwise.

There are some interests in RL to tackle those issues and to apply RL methods on real world tasks. Inverse RL lets an agent approximate a cost function from the demonstrations \cite{ng2000algorithms}, whereas Imitation Learning can imitate the behaviors from demonstrations in an end-to-end approach \cite{ho2016generative, hester2018deep}. Hierarchical RL decomposes a Markov decision process (MDP) into a hierarchy of smaller MDPs, together with the decomposition of corresponding value functions \cite{dietterich2000hierarchical}. Those algorithms are valuable for the real world applications of RL, while they are not irrelevant to traditional RL methods and still use algorithms developed in game environments.

An agent needs to balance the rate of exploration and exploitation in order to maximize the obtained rewards during a RL procedure. The experienced outcomes for an action at a certain state from the exploration phase can be preserved in a value-based manner, which are called Q-values.

There are two types of RL algorithms, distinguished by how they use the behaviour policy and target policy. On-policy algorithms evaluate and improve the same policy that is used to make moves. In other words, their behavior policy and target policy are the same. Q-learning is an off-policy RL method that adopts the different behaviour policy and target policy, which stores and updates Q-values based on the Bellman Optimality Equation by (Eq. \ref{eq:q}) \cite{watkins1992q}. Here, $\gamma$ is the discounting factor, $Q^*(s,a)$ is the Q-value at state $s$ with action $a$, $r_{t+1}$ is the result value obtained by advancing to state $s_{t+1}$ at time instant $t+1$. Q-learning is off-policy because $\mathrm{max}$ chooses a greedy action, but not necessarily the actual action the agent takes at $s_{t+1}$.

\scriptsize
\begin{equation}
    Q^*(s,a) = E\{r_{t+1}+\gamma \mathrm{max}_{a'}Q^*(s_{t+1},a'|s_t=s,a_t=a)\}
    \label{eq:q}
\end{equation}
\normalsize

\IEEEpubidadjcol

As RL continuously evolves, neural networks (NNs), including convolutional neural networks (CNNs) have been adopted to assist with the updating and exploring routines of RL. Deep reinforcement learning (DRL) is a combination of RL and deep learning (DL) architectures. Deep Q-learning Network (DQN) is one of the DRL methods that records Q-values in NNs instead of the traditional Q-tables. It was designed to play on the Atari games and achieved better than average human scores through feeding their CNN with raw image pixels of games as input \cite{mnih2013playing}.

One of the main problems of Q-learning is overestimation because the $\mathrm{max}$ operator is repeatedly used during the training. The $\mathrm{max}$ operator perpetuates the approximation errors in one direction, i.e., overestimation, but not underestimation. Overestimation bias can often contribute to suboptimal performances of Q-learning \cite{thrun1993issues} and lead the model to learn from repetitive moves that can destabilize the weights in the model. Double Q-learning \cite{hasselt2010double} is one of the most successful proposed methods to alleviate the overestimation bias problem in Q-learning. Double Q-learning maintains two separate Q-value functions but only updates one of them according to either (Eq. \ref{eq:QA}) or (Eq. \ref{eq:QB}). The learning rate is $\alpha(s,a)$, and $a^*$ is $\mathrm{argmax}_aQ_\theta^A(s',a)$, and $b^*$ is $\mathrm{argmax}_aQ_\theta^B(s',a)$ at the next state $s'$. The complete proof of convergence of Double Q-learning under similar conditions of Q-learning is given by \cite{hasselt2010double}, in their Theorem 1.

\scriptsize
\begin{equation}
    \label{eq:QA}
    Q^A(s,a) = Q^A(s,a) + \alpha(s,a)(r+\gamma Q^B(s',a^*)-Q^A(s,a))
\end{equation}
\normalsize
\scriptsize
\begin{equation}
    \label{eq:QB}
    Q^B(s,a) = Q^B(s,a) + \alpha(s,a)(r+\gamma Q^A(s',b^*)-Q^B(s,a))
\end{equation}
\normalsize

Double Q-learning is found to be unequivocally useful when it is integrated with NNs in DRLs. It has become a widely used technique in DRL implementations. Moreover, the convention is that the target network is kept as a copy of the policy Q-network instead of as a separate network in practice, designed to reduce the amount of computation overhead \cite{van2016deep}. The amount of steps to synchronize those two NNs is a hyper-parameter to be tuned.

In the meantime, Double Q-learning introduces a new problem called underestimation. Underestimation occurs because the target approximation is a weighted estimate of unbiased expected values, which are lower or equal to the maximum expected values \cite{hasselt2010double}. Although underestimation is often considered harmless in most algorithms, it may still introduce certain undesired effects \cite{chen2021investigation}.

Similarly, Bootstrapped Deep Q-Learning Network (Bootstrapped DQN) \cite{osband2016deep} is one of the methods that aspires to boost the performance of DQN through some simple but efficient modifications. Bootstrapped DQN intends to ameliorate the limited exploration capabilities of Q-learning by enabling deep exploration in complex RL tasks. Randomized value functions methods which can be implemented similarly to Thompson sampling \cite{thompson1933likelihood}, \cite{russo2017tutorial}, \cite{osband2015bootstrapped}, \cite{osband2016generalization} share some common traits with Bootstrapped DQN in regard to uncertainty estimation. Nevertheless, previous deep exploration methods either only work on MDPs with small finite state spaces, or relies on computationally intractable planning \cite{guez2012efficient, jaksch2010near}.

The importance of diversity in RL has been stressed and discussed in \cite{eysenbach2018diversity}, where the author replaces the reward function with a surrogate reward which is based on the entropy and mutual information terms. Each skill is sampled from a prior distribution $p(z)$ and should select trajectories as diversified as possible after the training. Their methods have shown promises in solving simple motor control tasks and indicates that diversity can be an important consideration when designing RL algorithms.

Previous research has also explored the effects of noise on RL \cite{plappert2017parameter,fortunato2017noisy,osband2019deep,touati2020randomized}, and in particular, a method that introduced a prior network to improve the diversity of Bootstrapped DQN \cite{osband2018randomized}, where an interesting way of integrating priors to Bootstrapped DQN was through simple additions of an untrainable prior network to each ensemble member. However, it was observed afterwards that this way of adding priors to Bootstrapped DQN does not improve the performance over Bootstrapped DQN \cite{touati2020randomized}. 

There are also many other techniques and algorithms that improve the performance of DQN,  such as Dueling Q-learning \cite{wang2016dueling}. However, they are orthogonal to the original study of Bootstrapped DQN.

\section{Method}
In DQN, there is typically only one output head, but Bootstrapped DQN consists of multiple ($K$) output heads. This major modification allows Bootstrapped DQN to make actions based on a particular head during exploration, yet to utilize an ensemble of votes during evaluation. A randomly pre-selected head is to be used for each game episode in the exploration phase of Bootstrapped DQN. During the updating phase, there are head masks generated from a Bernoulli distribution to decide whether or not the heads should be updated given the currently drawn replay sample. Bernoulli probability is a value in [0,1] to determine how often 1 is drawn from \{0,1\}. Increasing the value of Bernoulli probability equates to more heads being trained on the same batches of samples, which in turn reduces the diversity of Bootstrapped DQN. Surprisingly, a large value for the Bernoulli probability would not degenerate the performance of the algorithm. The masks were found out best to be 1 or a value close to 1, which can accelerate the training speed with little influences on performances compared to when they are set as smaller values.

The benefit of keeping multiple heads is that there is a diversity of network weights among different heads. In DRL implementations, NNs are typically trained more than millions of times in the learning process, in which identical samples can be drawn repetitively from the replay memory in spite of their distinguished individual Bernoulli probability values in Bootstrapped DQN. As a drawback of this type of intensive training, smaller values of masks in Bootstrapped DQN not only cannot improve the performance of the model, but also decelerates the training and the convergence \cite{osband2016deep}. It is also obvious that the diversity introduced by using $K$ heads will eventually be negligible after they are trained on the same subset of data frequently and there is no new trajectories to be successfully discovered.

Bootstrapped DQN treats the randomly initialized network weights as a prior. However, this random prior cannot assure the diversity after a decent amount of updates. Bootstrapped DQN also uses a shared network body to learn a joint feature representation, so that the predictions of all heads will eventually become similar or even the same to each other. In the search space that has already been visited thoroughly, different heads would be more likely to converge into the same policy and thus the rate of exploration will be low \cite{ecoffet2019go}. A high level of diversity amongst heads is only guaranteed at the states that have never been encountered or been rarely drawn from the replay memory.

Combining this interpretation with the observation from \cite{touati2020randomized}, the same conclusion can be drawn that the Bootstrapped DQN plus Prior method from \cite{osband2018randomized} still suffers from the same degraded diversity problem at the late stage where the model has been trained many times. As a side-effect of using fixed randomized priors, the benefit of diversity only exists in the states that have been trained infrequently, and then will be overshadowed by the tremendous amount of updates.

Although we agree on the principles of using randomized priors in \cite{osband2018randomized}, we argue that a successful method to combine priors should satisfy the following requirements:

\begin{enumerate}
\item Priors have to be independently drawn each time of the NN update, to maintain the level of diversity at the late stage of training.
\item The method should not strengthen the overestimation or the underestimation problem, in other words, not collapsing the model.
\item Priors should still play a role significantly enough to result in more efficient exploration and optimum convergence despite the above point.
\item The method to obtain priors should not be computationally costly.
\end{enumerate}

Thus, our hypothesis is that noise is more suited than a fixed prior in order to satisfy all those requirements. We propose our algorithm based on Bootstrapped DQN, which removes the role of priors and introduces more diversity into the algorithm. Replacing priors with noise can introduce diversity by slightly shifting the target value even on a state that has already been visited frequently. In turn, they expand the search horizon of Bootstrapped DQN and can help avoid being stuck in a suboptimal solution for a long period at the late training stage.

Our proposal of modification is to combine the noise with target values during the network update. In each update, a distinct noise value is uniquely sampled, scaled and then added to the target value. This modification requires less adjustments from Bootstrapped DQN comparing to changing the parameters of the NN instead.

In order to minimize the amount of computations needed to generate noise at each update, we choose to use a Gaussian distribution $G(\mu,\sigma)$ for sampling noise, which is considerably computationally more efficient than generating a noise network in each update or changing the structure of NNs. Directly operating noise on target values also simplifies the path to find a suitable scaling parameter because the Q-values themselves can serve as anchor points. A larger noise value is desired as the Q-values grow along the training process. The absolute values of noise are therefore scaled in accordance with the actual Q values to be compatible with games having large score values. The parameter $scale$ defined in (Eq. \ref{eq:scale}) linearly increases our noise values. Here, $\beta$ is a small positive parameter.

\scriptsize
\begin{equation}
scale = 1 + \beta*\mathrm{max}_a Q_{\theta}^A (s, a)
    \label{eq:scale}
\end{equation}
\normalsize

From this equation, it can be seen that $scale$ approximates 1 when $\beta*\mathrm{max}_a Q_{\theta}^A (s, a)$ is small enough but remains positive. Although there is a risk that the algorithm might exhibit some unstable behaviors at the beginning of training due to small target Q-values oscillating by relatively large noise values, we still consider as necessary to have such a lower bound of the noise. As mentioned above, Double Q-learning has the underestimation problem and could potentially yield extremely low Q-value predictions even at the late phase of training. A linear noise without a lower bound might scale to a value that is too small to be any significant in this case. Hence, a lower bound of the noise helps Bootstrapped DQN escape the local optimum and explore more efficiently, whereas the initial dithering can be obscured by a large $\epsilon$ at the beginning of training where the agent mostly chooses random actions.

The Q-values are typically lower than the maximum expected value due to the underestimation of Double Q-learning. However, this underestimating behaviour is susceptible to the influences of the choice of $\beta$ in $scale$. A Gaussian distribution that only generates negative values could oppress the exploring behavior of the agent and hence leads to bad convergence. On the contrary, the underestimating could potentially turn into overestimating if the distribution coincidentally produces any positive values because the intensive NN updates in RL can lead to overestimating even with a trivial positive bias. Thus, $\beta$ should be designed as a relatively small (positive) value in our algorithm.

Considering the case where $\mathrm{max}_a Q_{\theta}^A (s, a)$ is non-positive and then $scale\leq1$, larger $\mathrm{max}_a Q_{\theta}^A (s, a)$ would still result in a larger $scale$ as it previously does. Thus, there is no special treatment required for non-positive Q-values.

Our pseudo-code is shown in Algorithm \textbf{\ref{algo:mine}}. In our algorithm, $r$ is the reward, $\gamma$ is the discounting factor, $\epsilon$ is the exploration ratio, $maxFrames$ is the maximal number of total frames, and $L()$ is the loss function, i.e., Smooth L1 loss in this case. $M$ is a Bernoulli distribution and $G$ is a Gaussian distribution, from which we draw $m$ and $np$ respectively. K is the number of network heads.

We have two neural networks. $Q_\theta^A$ is the policy network and $Q_\theta^B$ is the target network. $Q_\theta^B$ is a duplicate of $Q_\theta^A$ and they synchronize at an interval of $sync$ frames. For the update, $a^*$ is $\mathrm{argmax}_aQ_{\theta k}^A(s',a)$, $s$ is the current state, $s'$ is the next state, $a$ is the action, $terminal$ is a flag indicating whether the game ends or not.

\begin{algorithm}[tb]
\scriptsize
\caption{Bootstrapped DQN with Priors Replaced by Noise}\label{algo:mine}
\textbf{Input}: $\epsilon$, $\gamma$, $maxFrames$, $sync$, $M$, $L()$, $G$, $K$\\
\textbf{Parameter}: D, $Q_\theta^A$,$Q_\theta^B$\\
\textbf{Output}: $Q_\theta^A$
\begin{algorithmic}[1]
\STATE $frames \gets 0$
\WHILE{$frames<maxFrames$}
\STATE Initialize the environment
\STATE Pick a head $k$ uniformly from $\{0, ..., K-1\}$ to make actions

\WHILE{Game not finished}
\STATE $frames$ += $4$
\IF{$random()<\epsilon$}
\STATE Choose action randomly
\ELSE
\STATE Choose action based on $Q_{\theta k}^A$
\ENDIF
\STATE Store $s,a,s', r, terminal$ into replay buffer D
\STATE Store sampled random masks $m$ from M into replay buffer D
\STATE Draw minibatch of $s,a,s', r, terminal, m$ from D
\STATE Update $Q_\theta^A$ according to Algorithm \ref{algo:mine2}
\ENDWHILE
\IF {$frames\%sync==0$}
\STATE $Q_\theta^B \gets Q_\theta^A$
\ENDIF
\ENDWHILE
\STATE \textbf{return} $Q_\theta^A$
\end{algorithmic}
\end{algorithm}
\normalsize

\begin{algorithm}[tb]
\scriptsize
\caption{Network Update}\label{algo:mine2}
\textbf{Input}: $s,a,s', r, terminal, m$, $L()$, $G$, $K$\\
\textbf{Parameter}: $Q_\theta^A$,$Q_\theta^B$, $scale$\\
\textbf{Output}:
\begin{algorithmic}[1]
\STATE Generate noise $np$ from the Gaussian distribution $G$
\STATE $totalLoss=0$
\FOR{$k$ in $\{0, ..., K-1\}$}
\STATE $target = r+\gamma Q^B_{\theta k}(s',a^*)*(1-terminal)+scale*np_k$
\STATE Compute $loss$ by $L(Q^A_{\theta k}(s,a), target)$
\STATE Mask $loss$ with $m$
\STATE $totalLoss$ += $loss$
\ENDFOR
\STATE $totalLoss$ /= $K$
\STATE Update $Q_\theta^A$ by $totalLoss$ with gradient descent
\STATE \textbf{return}
\end{algorithmic}
\end{algorithm}
\normalsize

\section{Experimental Details}

\begin{figure}[t]
    \centering
    \includegraphics[width=0.8\linewidth]{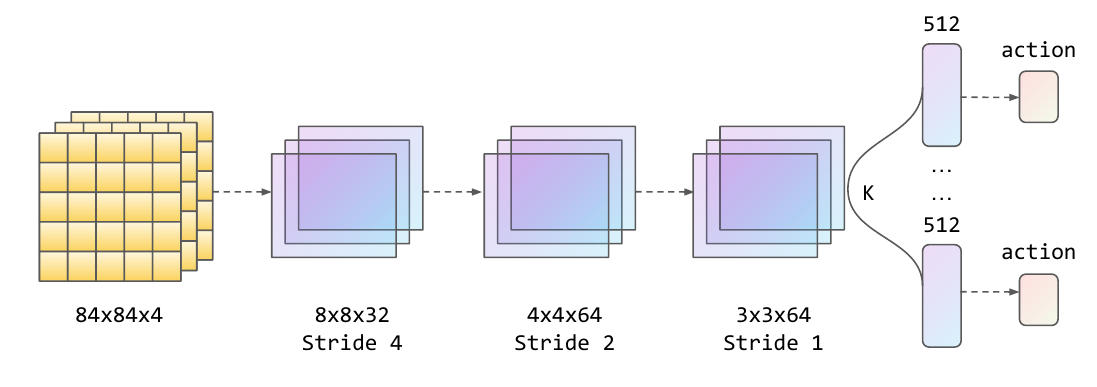}
    \caption{The structure of our NN. It takes the same input and have the same parameters of convolutional layers as in DQN, and then branches into K heads. Each head contains a fully connected layer, followed by an output layer.}
\label{fig:nn}
\end{figure}

Our experiments are conducted across 49 Atari games on Arcade Learning Environment (ALE), the same as Bootstrapped DQN \footnote{Code can be found in https://github.com/mengli11235/Bootstrapped-DQN-with-NP}. For a more detailed ablation study on the effects of using noise, we include evaluation curves of 10 games out of them. The hyper-parameters and NN architectures are also kept mostly the same to those stated in Bootstrapped DQN. The NN architecture is shown in Fig. \ref{fig:nn}. The list of chosen parameters is shown in Table \ref{tbl:para}.

Adam is an optimization algorithm that uses the adaptive estimates of lower order moments of the gradient \cite{kingma2014adam}, whereas RMSProp computes the momentum based on rescaled gradients. Although most hyper-parameters followed the chosen values of Bootstrapped DQN, we use Adam instead of RMSProp as our optimizer because Adam shows more stable performance on Bootstrapped DQN in our experiment. For similar reasons, the value of $K$ is set to 9 instead of 10 with the Bernoulli probability decreased from 1 to 0.9. The hyper-parameter choices of noise-related values are found out through grid search. However, different values of $\beta$ are found to have similar effects as long as it is set around $0.05$.

In most of the Atari games, small rewards were commonly designed to lead to an optimal policy. Thus, the rewards are clipped between -1 and 1 in our experiment to improve the stability. Meanwhile, the reported scores are still raw scores in our results section.

We also use a random no-operation (no-op) period of $[0,30)$ applied at the beginning of each game episode in order to introduce stochasticity into the environment. The seed of the Atari environment is fixed during the training phase, but randomly selected seeds are used for the evaluation process. We run 5 episodes with those random seeds for each instance of the evaluation phase.

\begin{table}[htbp]
\caption{Parameters used in our experiment}
\centering
\begin{tabular}{ |c||c|} 
 \hline
    Input & $84\times84\times4$ \\ 
  \hline
      K & 9 \\ 
  \hline
      Bernoulli probability & 0.9 \\ 
  \hline
      Optimizer & Adam \\ 
  \hline
        Adam learning rate & 0.0000625 \\ 
  \hline
      $\gamma$ & 0.99 \\ 
  \hline
      Initial $\epsilon$ & 1 \\
  \hline
      Final $\epsilon$ & 0.01 \\
  \hline
      $\epsilon$ decay frames & 1M \\
  \hline
      $sync$ & 40000 \\
    \hline
      Frames per step & 4 \\ 
  \hline
      Steps per evaluation & 250000 \\ 
  \hline
      MaxFrames & 200M \\ 
  \hline
      Replay size & 1M\\
  \hline
      Batch size & 32\\
  \hline
      Noise $\mu$ & 0 \\
  \hline
      Noise $\sigma$ & 0.02 \\
  \hline
      Noise $\beta$ & 0.05 \\
  \hline
\end{tabular}
\label{tbl:para}
\end{table}

\section{Results}
\begin{figure}[t]
    \centering
    \includegraphics[width=0.8\linewidth]{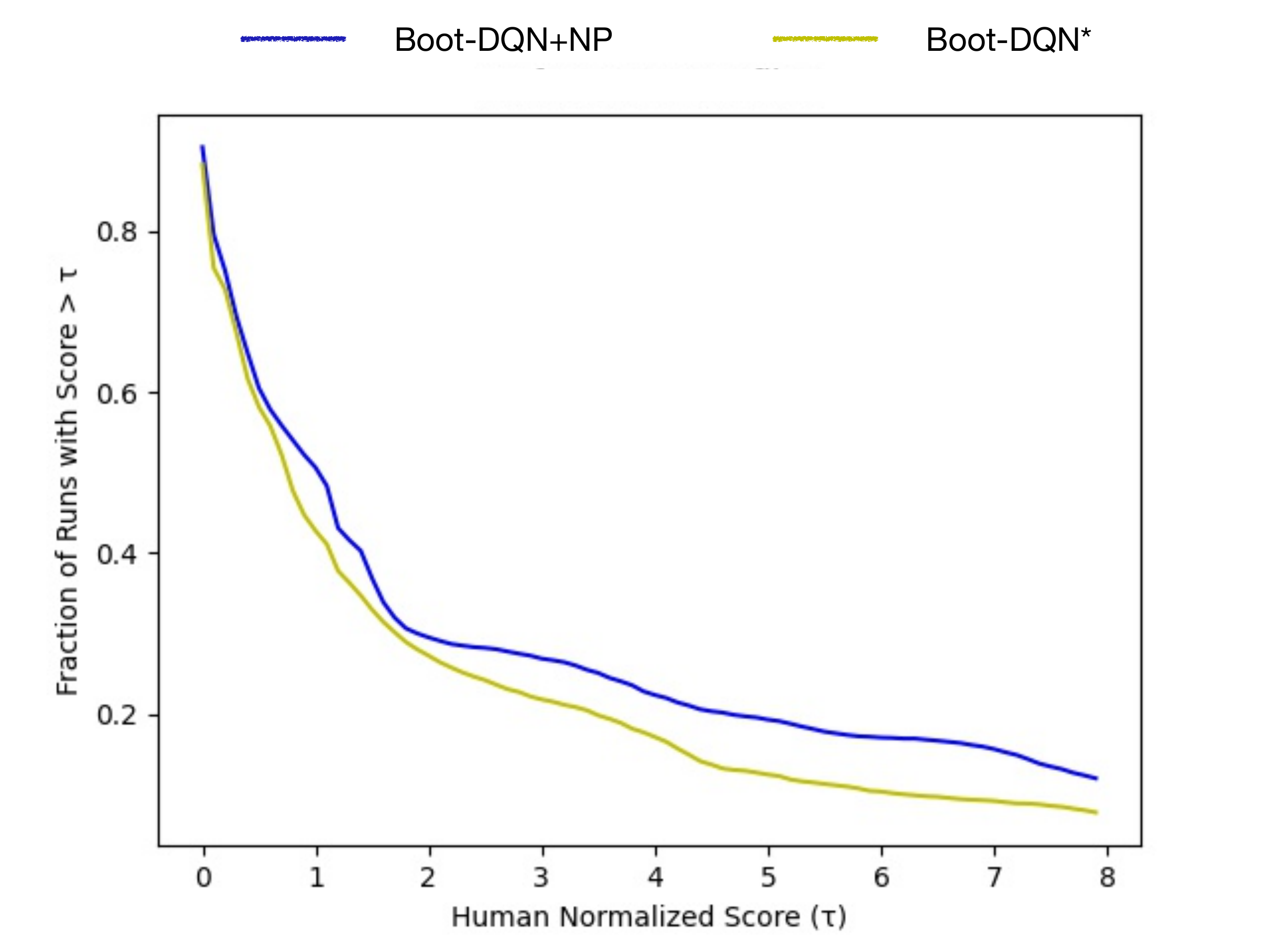}
    \caption{The performance profiles of 49 Atari games up to $\tau=8$. The blue line is the percentage of the score larger than $\tau$ using noise. The orange line is the percentage of that without using noise.}
\label{fig:perf1}
\end{figure}

\begin{figure*}[t]
    \centering
    \includegraphics[width=0.8\linewidth]{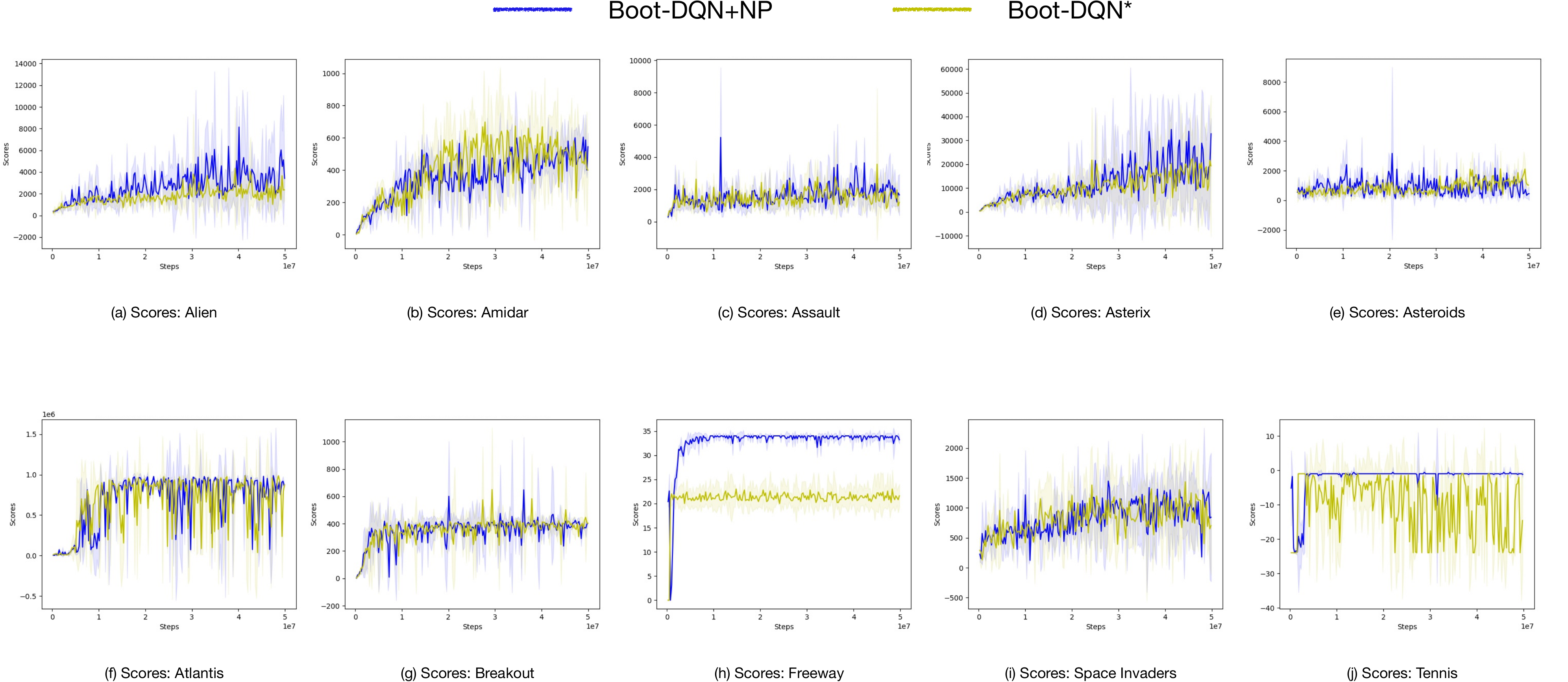}
    \caption{The evaluation results of playing 10 Atari games. The blue lines are the mean evaluation scores of using noise. The orange lines are the mean evaluation scores without using noise. The 95\% confidence intervals are also plotted in addition to the scores. All are evaluated every 250000 steps.}
\label{fig:res1}
\end{figure*}

We present our evaluation curves in terms of means and standard deviations in Fig. \ref{fig:res1} to show some detailed results on 10 games. We choose those games both by alphabets and by representativeness. The first 6 games are selected alphabetically, and the rest 4 are well-known classic Atari games. As the original paper of Bootstrapped DQN provided full maximal evaluation scores on 49 Atari games, we present comparable full results of maximal evaluation scores and include them in Table \ref{tbl:res1}. To provide more advanced analysis over aggregated data, the measurement of performance profiles (score distributions) described in \cite{agarwal2021deep} with overall human normalized scores is shown Fig. \ref{fig:perf1}.

To concisely represent contents in figures and tables, we use \textit{'Boot-DQN'} as an abbreviation for the Bootstrapped DQN algorithm in the original paper, \textit{'Boot-DQN*'} for our Bootstrapped DQN implementation without using noise, and \textit{'Boot-DQN+NP'} for Bootstrapped DQN with priors replaced by noise. The only difference between \textit{'Boot-DQN+NP'} and \textit{'Boot-DQN*'} is whether they use the noise (our algorithm) or not (the default algorithm).

In Fig. \ref{fig:res1}, the mean evaluation scores and the 95\% confidence intervals of \textit{'Boot-DQN+NP'} and \textit{'Boot-DQN*'} across different games are being compared. The 95\% confidence intervals are obtained from 5 evaluation episodes for each step of a single training run. Oscillations of evaluation scores in both algorithms can be attributed to the different game initializations due to random seeds in the evaluation phase, and the stochasticity we introduce through no-op frames.

It is clear that our algorithm achieves not only higher final mean scores, but also high mean evaluation scores during most of the training period than the default algorithm across the majority of games. In fact, the only game that \textit{'Boot-DQN*'} shows significantly better performance than \textit{'Boot-DQN+NP'} is Amidar. \textit{'Boot-DQN+NP'} beats \textit{'Boot-DQN*'} in almost all other games in terms of mean scores and confidence intervals.

At the beginning of each game, the differences of evaluation scores between these two algorithms are usually not significant. The default algorithm without using noise can even obtain better scores among some of the games, which can be attributed to the dithering effects of setting a lower boundary for $scales$, as described in the method section.

Occasionally, the default algorithm plateaus after approximately 1 million steps of training and its performances cannot improve meaningfully till the end of training. In those games, our algorithm with noise is less likely to find itself in a similar plateau, and even if it gets stuck, it is then able to demonstrate an entirely new behavior -- breaking the plateau and achieving far higher scores at the end of training. Notably, our algorithm seems to be able to continue to keep increasing scores after 200 million frames in more games, given its better ability of escaping the plateau, which can be another promising feature.

In most cases, this 'plateau' is not stationary and could accompany heavy drops in scores later in the training, due to the varied game difficulties along the episode caused by versatile game mechanisms. Both algorithms suffer from that and they may or may not recover from those downhills. Nevertheless, it is important to notice that \textit{'Boot-DQN+NP'} vastly obtains higher mean evaluation scores with confidence intervals than \textit{'Boot-DQN*'} in most cases where this happens. Moreover, \textit{'Boot-DQN+NP'} often bests \textit{'Boot-DQN*'} in final scores at the end of training provided they are able to recover from the downhill.

Meanwhile, there are some games in which there is no obvious plateau at all. There seems to be constant improvements for both algorithms throughout the whole training process. Even so, \textit{'Boot-DQN+NP'} shows faster convergence and demonstrates better performance than \textit{'Boot-DQN*'}. This entails that noise can facilitate the environment exploration and finds improved trajectories faster. This is especially true at the late training stage when $\epsilon$-greedy is almost not at play, despite the fact that noise is not designed to replace $\epsilon$-greedy at the first place.

\begin{figure*}[t]
    \centering
    \includegraphics[width=0.8\linewidth]{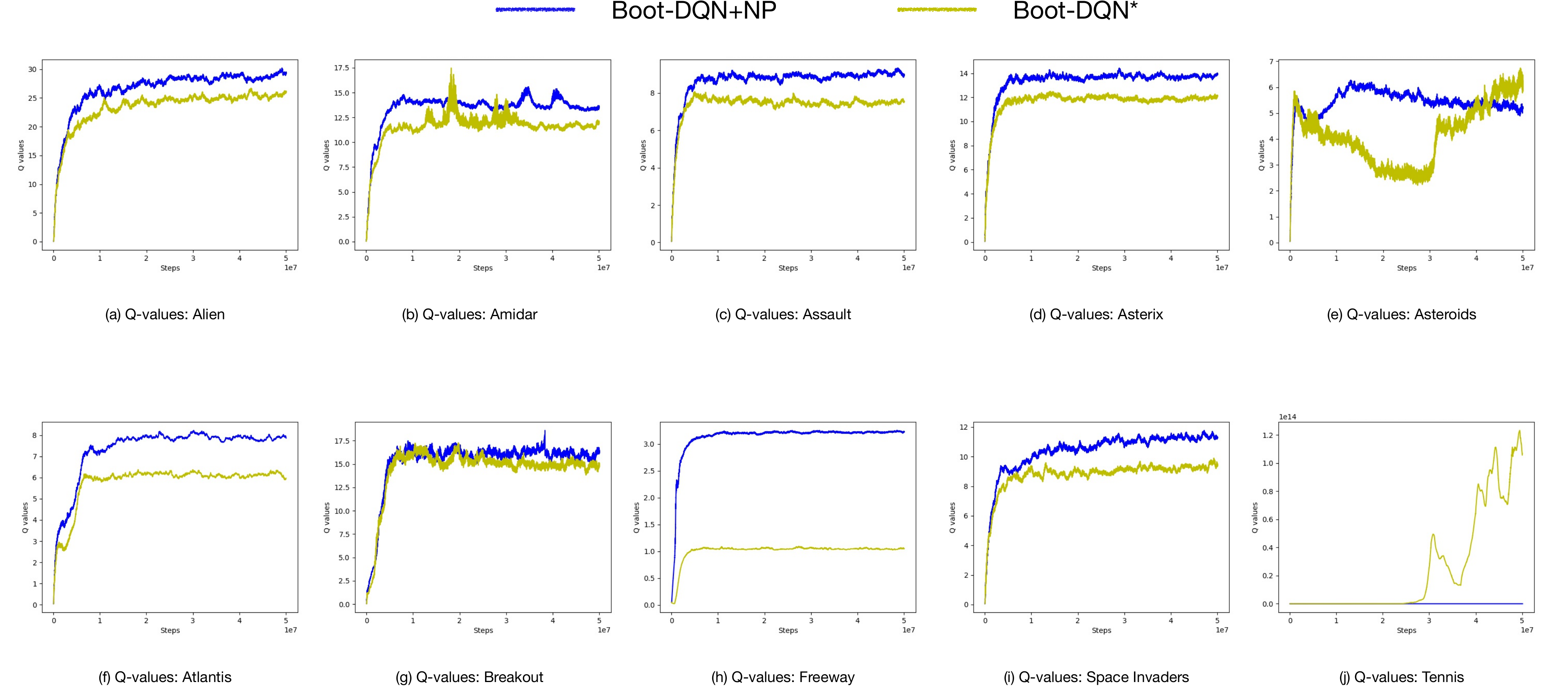}
    \caption{The Maximal Q-values corresponding to Fig. \ref{fig:res1}. Maximal Q-values are updated each time the network gets updated. Q-values are from the predictions of the policy network. Those values are used to update the parameter $scale$.}
\label{fig:res2}
\end{figure*}

In Fig. \ref{fig:res2}, we show the maximal Q-values predicted by the policy network alongside the period of training. It can be seen that adding noise increases the maximal Q-value predictions, but not to an extent where it crashes the model by overestimation. In fact, the Q-values are still underestimated in general, and the influences of adding noise are more salient in games where severe underestimation occurs. In the game of Asteroids, where \textit{'Boot-DQN*'} shows unstable behaviors and inflated Q-values, the maximal Q-values of \textit{'Boot-DQN+NP'} surprisingly remain stable and still maintain smaller at the end. In Tennis, the maximal Q-values of \textit{'Boot-DQN*'} overestimates to an absurdly large value, whereas those of \textit{'Boot-DQN+NP'} still remain at reasonable values. Results from both games demonstrate that adding noise enhances the stability of our algorithm and \textit{'Boot-DQN+NP'} even possesses the ability to achieve better performance than \textit{'Boot-DQN*'} due to this improvement. This improvement is visible up to 200M frames in tennis, but does not remain till the end in Asteroids. Although it is not always the case that this improvement can be sustainable throughout the whole training process, it still results in higher maximal evaluation scores. In our view, adding bias to the model in this way serves as an additional regularizer and can be an excellent example of beneficial variance-bias trade-off.

In Table \ref{tbl:res1}, the results for \textit{'Boot-DQN'} and \textit{'DDQN'} are taken from \cite{osband2016deep}. \textit{'DDQN'} is their improved implementation of the DQN with their own specified parameters. In contrast, \textit{Nature} is the original DQN implementation from \cite{mnih2015human}, which performs slightly worse than \textit{'DDQN'}. It might appear that \textit{'Boot-DQN*'} obtains better results than the original \textit{'Boot-DQN'} by merely using different parameters in various games. In some other games, however, there are degraded performances for \textit{'Boot-DQN*'} or no significant differences between the two of them. In Freeway, the maximal evaluation score of \textit{'Boot-DQN*'} drops tremendously from 34 to around 26. Moreover, it is likely that the original \textit{'Boot-DQN'} used averages of maximal scores, either by rolling averages or by multiple runs. Thus, it is important to direct our focus towards how using noise improves the performance of Bootstrapped DQN (i.e., difference between \textit{'Boot-DQN*'} and \textit{'Boot-DQN+NP'}), but not towards parameter tunings and calibrations.

We highlight the algorithm which has achieved the best scores. Table \ref{tbl:res1} clearly shows that \textit{'Boot-DQN+NP'} achieves better results compared to both \textit{'Boot-DQN'} and \textit{'Boot-DQN*'} in terms of the maximal evaluation scores. The benefit of introducing noise alone boosts the performances of the Bootstrapped DQN algorithm tremendously. In fact, the performance of scores might be even larger if by checking some of the intermediate results in Fig. \ref{fig:res1}. The maximal evaluation scores for our algorithm are higher or equal than those of \textit{'Boot-DQN*'} in 30 out of 49 (61.2\%) games.

Performance profiles give an aggregated score distribution on all tasks as a whole that allows qualitative comparisons. In Fig. \ref{fig:perf1}, it is clear that \textit{'Boot-DQN+NP'} has similar values as \textit{'Boot-DQN*'} at the beginning, but significantly higher values approximately after $\tau=2$. \textit{'Boot-DQN+NP'} having a larger tail distribution entails an overall improved superhuman performance the algorithm is capable of achieving when $\tau>2$.

\begin{table*}[htbp]
\caption{Full Results by Maximal Evaluation Scores}
\centering
\begin{tabular}{ |c||c|c c|c c|c|c|} 
 \hline
    \textbf{Game} & \textbf{Boot-DQN} & \textbf{Boot-DQN*} &\textbf{(Normalized)} &\textbf{Boot-DQN+NP} &\textbf{(Normalized)} & \textbf{DDQN} & \textbf{Nature}\\
  \hline
  Alien & 2436.6 & 7700 &(1.083) & \textbf{12330}&(1.754) & 4007.7  & 3069\\
  \hline
    Amidar & 1272.5 & 873 &(0.506)& 728 &(0.421)& \textbf{2138.3} & 739.5\\
  \hline
    Assault & 8047.1 & 8044&(15.053) & \textbf{8223} &(15.476)& 6997.9 & 3359\\
    \hline
    Asterix & 19713.2 & 40900 &(4.906)& \textbf{46500} &(5.582) & 17366.4 & 6012\\
    \hline
    Asteroids & 1032 & 3320&(0.054) & \textbf{8960}&(0.177)& 1984.4 &1629\\
    \hline
    Atlantis & 994500 & 1000700&(61.061) & \textbf{1005900}&(61.382) & 767850 & 85641\\
    \hline
    Bank Heist & 1208 & \textbf{1300}&(1.74) & \textbf{1300}&(1.74) & 1109 & 429.7\\
    \hline
    Battle Zone & 38666.7 & \textbf{61000}&(1.684) & \textbf{61000}&(1.684) & 34620.7 & 26300\\
    \hline
    Beam Rider & 23429.8 & 28850&(1.72) & \textbf{34936}&(2.087) & 16650.7 & 6846\\
    \hline
    Bowling & 60.2 & 30 &(0.05)& \textbf{78}&(0.4) & 77.9 & 42.4\\
    \hline
    Boxing & 93.2 & \textbf{100} &(8.325) & \textbf{100} &(8.325)& 90.2 & 71.8\\
    \hline
    Breakout & \textbf{855} & 834 &(28.899) & 822 &(28.483) & 437 &401.2\\
    \hline
    Centipede & 4553.5 & \textbf{13980}&(1.2) & 12383&(1.037) & 4855.4 & 8309\\
    \hline
    Chopper Command & 4100 & 3200&(0.363) & 2400 &(0.242) & 5019 & \textbf{6687}\\
    \hline
    Crazy Climber & 137925.9 & 195000&(7.354) & \textbf{208000}&(7.873) & 137244.4 & 114103\\
    \hline
    Demon Attack & 82610 & 28475 &(15.571) & 35255&(19.3) & \textbf{98450} & 9711\\
    \hline
    Double Dunk & 3 & \textbf{4}&(10.273) & 2 &(9.364)& -1.8 & -18.1\\
    \hline
    Enduro & 1591 & 2286 &(2.657)& \textbf{2364} &(2.747)& 1496.7 & 301.8\\
    \hline
    Fishing Derby & 26 & \textbf{41} &(2.504) & 39 &(2.466)& 19.8 & -0.8\\
    \hline
    Freeway & 33.9 & 26 &(0.878)& \textbf{34} &(1.149)& 33.4 & 30.3\\
    \hline
    Frostbite & 2181.4 & \textbf{4690}&(1.083) & 2430&(0.554) & 2766.8 & 328.3\\
    \hline
    Gopher & 17438.4 & \textbf{118380}&(54.816) & 118340&(54.797) & 13815.9 & 8520\\
    \hline
    Gravitar & 286.1 & \textbf{2000}&(0.575) & 1050&(0.276) & 708.6 & 306.7\\
    \hline
    Hero & \textbf{21021.3} & 13415&(0.416) & 13375 &(0.414)& 20974.2 & 19950\\
    \hline
    Ice Hockey & -1.3 & 2 &(1.091) & \textbf{3} &(1.174)& -1.7 & -1.6\\
    \hline
    Jamesbond & 1663.5 & \textbf{7600}&(27.652) & 7000 &(25.46) & 1120.2 & 576.7\\
    \hline
    Kangaroo & 14862.5 & 14500 &(4.843)& \textbf{16600}&(5.547) & 14717.6 & 6740\\
    \hline
    Krull & 8627.9 & \textbf{11430} &(9.21)& 10764&(8.587) & 9690.9 & 3805\\
    \hline
    Kung Fu Master & 36733.3 & 52800 &(2.337) & \textbf{56900}&(2.52) & 36365.7 & 23270\\
    \hline
   Montezuma Revenge & \textbf{100} & 0 &(0) & 0&(0) & 0 & 0\\
    \hline
   Ms Pacman & 2983.3 & 5620 &(0.8)& \textbf{5970} &(0.852) & 3424.6 & 2311\\
    \hline
   Name This Game & 11501.1 & 13950 &(2.025)& \textbf{15180}&(2.239) & 11744.4 & 7257\\
    \hline
   Pong & 20.9 & \textbf{21} &(1.181)& \textbf{21} &(1.181)& 20.9 & 18.9\\
    \hline
   Private Eye & 1812.5 & \textbf{15100}&(0.217) & 4000 &(0.057)& 158.4 & 1788\\
    \hline
   Qbert & 15092.7 & 26375 &(1.972)& \textbf{27600} &(2.064) & 15209.7  & 10596\\
    \hline
   Riverraid & 12845 & 15760 &(0.914)& \textbf{17320}&(1.013) & 14555.1 & 8316\\
    \hline
   Road Runner & 51500 & 70500&(8.998) & \textbf{82100}&(10.479) & 49518.4 & 18257\\
    \hline
   Robotank & 66.6 & 54 &(5.34)& 52&(5.134) & \textbf{70.6} & 51.6\\
    \hline
   Seaquest & 9083.1 & 28830 &(0.555)& \textbf{30800} &(0.732)& 19183.9 & 5286\\
    \hline
   Space Invaders & 2893 & 1950 &(1.185)& 2100 &(1.284)& \textbf{4715.8} & 1976\\ 
    \hline
   Star Gunner & 55725 & 55300 &(5.7)& 60900&(6.284) & \textbf{66091.2} & 57997\\
    \hline
   Tennis & 0 & 1&(1.6) & 1 &(1.6)& \textbf{11.8} & -2.5\\
  \hline
   Time Pilot & 9079.4 & \textbf{15700}&(7.303) & 8600 &(3.029)& 10075.8 & 5947 \\
    \hline
   Tutankham & 214.8 & 391 &(2.43)& \textbf{399} &(2.481)& 268 & 186.7\\
    \hline
   Up N Down & 26231 & \textbf{70120}&(6.235) & 56380&(5.004) & 19743.5 & 8456\\
    \hline
   Venture & 212.5 & \textbf{1700} &(1.432)& 0 &(0)& 239.7 & 380\\
    \hline
   Video Pinball & 811610 & \textbf{999004} &(56.543)& 894533 &(50.63)& 685911 & 42684 \\
    \hline
   Wizard of Wor & 6804.7 & 18700 &(4.325)& \textbf{24500}&(4.683) & 7655.7 & 3393\\
    \hline
   Zaxxon & 11491.7 & \textbf{17200}&(1.878) & 14100 & (1.539) &12947.6& 4977\\
    \hline
\end{tabular}
\label{tbl:res1}
\end{table*}

\section{Discussion}
The results show that our algorithm outperforms both the original algorithm and the algorithm without using noise in most of the games. However, it is not the case that this performance enhancement can occur amongst all games. In ALE, there are a variety of different types of games, with distinct game mechanisms, such as goal and sub-goal designs, reward assignments, agent and object controls. Those all imply that model-free RL methods without specific tuning or domain knowledge for each game can exhibit superhuman performances in a set of games, but not in others. Nonetheless, it is of the interest of AI researchers to avoid using specific tuning or domain knowledge per game, and to let an algorithm learn to solve various tasks at the human or superhuman level. Naturally, our algorithm can perform better in many games but not in certain ones. What is crucial is that we have exactly demonstrated that \textit{Boot-DQN+NP} yields improved performance than the default algorithm without using noise in the majority of the game results.

Our hypothesis indicates that such improvement on the performances of \textit{Boot-DQN+NP} originates from its expanded search horizon. \textit{Boot-DQN+NP} is capable of shifting the current spotlight from endlessly searching the attended trajectory to exploring other trajectories with seemingly lower rewards. However, there is no guarantee that the agent would direct its search efficiently towards the global optimum even if this optimum is within its expanded horizon. If the agent was able to return to a state with high rewards and promising discoveries first and then explore by some efficient strategies, its scores could potentially increase drastically. Thus, adding a policy-based Go-Explore mechanism is expected to further bolster our results \cite{ecoffet2021first}.

In Fig. \ref{fig:res1}, both agents find themselves in plateaus without reaching the possible maximal scores in a variety of games. They can escape this plateau sometimes, but it often requires a large amount of training episodes. Ideally, one might wish to tune $\beta$ a bit higher in order to reduce the amount of the period in which the agent gets stuck on the plateau. However, increasing the parameter $\beta$ slightly in our experiment would not demonstrate more significant improvement on the capability of escaping plateaus, and increasing it too much would cause the model to crash due to overestimation. Besides, a complex game often demands a sufficient amount of explorations in order for the agent to learn new behaviors in the game.

We distinguish our algorithm from the famous NoisyNet \cite{fortunato2017noisy} in several ways. First, NoisyNet was proposed to replace the conventional exploration schemes such as $\epsilon$-greedy and entropy reward, whereas our algorithm does not replace those exploration strategies. We still use $\epsilon$-greedy to guide our explorations phase. Moreover, NoisyNet utilizes a parametric function to add noise into the weight and bias of their NN. It directly modifies the network architecture and introduces additional trainable parameters. On the other hand, our algorithm aims at improving the diversity of Bootstrapped DQN. It combines the noise with target values, which does not require any modifications to the network architecture, and hence not introduce any additional computational overhead on the network level.

In short, NoisyNet works on a more general RL topic and tries to find replacements for exploration schemes at the costs of more algorithmic complexity, whereas our method improves on a specific algorithm, Bootstrapped DQN, with adding minimal computations.

\section{Conclusion}
In this paper, we try to mitigate a problem common in Bootstrapped DQN, where the diversity degrades after a tremendous amount of training. A fixed prior network has been proposed to mediate this problem. Nonetheless, the integrity of diversity still cannot be maintained in the late training phase with this modification despite being eased slightly. Our algorithm introduces randomness into the diversity constantly of Bootstrapped DQN through noise. Priors here are being replaced by noise, with minimal extra computational overhead.

Our results demonstrate that our adjustment of the algorithm contributes to the improvements of evaluation scores in most of the 49 games we have trained and tested. Thus, our hypothesis has been substantiated and the benefits of utilizing noise are significant.

Meanwhile, it has been pointed out that sticky actions can reliably introduce more stochasticity in Atari games \cite{machado2018revisiting}. It is beneficial to implement sticky actions in further experiments.

Besides using noise, there are other interesting concepts such as applying self-supervised learning to RL, which aspires to building world models of background knowledge. It could potentially help the agent plan and explore the environment \cite{sekar2020planning}, \cite{hendrycks2019using}. Bootstrapped DQN utilizes the random initialization of network priors to perform random exploration on rarely trained states, but it certainly could benefit from a world model. The combination of using noise and a world model could be an interesting future work.

Decision Transformer has also shown the potential of solving RL tasks efficiently offline \cite{chen2021decision}, which can be viewed as a successful combination of self-supervised learning and RL. Transformers, in a broad sense, share some similar concepts with Bootstrapped DQN, because they both use multiple network heads and also masking/gating mechanisms to direct the diversity or the attention of their networks. Although Bootstrapped DQN found out that masking in practice cannot improve the performances but decreases the training speed, it might be interesting to adjust the mechanisms in accordance with those of transformers and to explore the undiscovered effects of masking more in the future.

Self-supervised learning and more specifically, transformers, can be a potential future research direction to further improve the performance of Bootstrapped DQN with noise. On the other hand, concepts from bootstrapped DQN with noise might potentially benefit the research of self-supervised learning as well.

\section*{Acknowledgments}
The research presented in this paper has benefited from the Experimental Infrastructure for Exploration of Exascale Computing (eX3), which is financially supported by the Research Council of Norway under contract 270053. We want to acknowledge the help received from the Department for Research Computing at USIT,
the University of Oslo IT-department. This work was performed on the [ML node] resource, owned by the University of Oslo, and operated by the Department for Research Computing at USIT,
the University of Oslo IT-department. http://www.hpc.uio.no/

\appendix[Evaluation Results after 200M Frames]
Mean scores, standard deviations and human normalized scores after 200M frames of training are reported in Table \ref{tbl:appen1}, which is a supplement of Fig. \ref{fig:res1}. In general, the mean scores are considerably lower than the maximal evaluation scores demonstrated in Table \ref{tbl:res1}, with large standard deviations, accounted both by the small number (5) of evaluation runs and by the stochasticity. \textit{Boot-DQN+NP} scores equal or higher than \textit{Boot-DQN*} in 33 out of 49 (67.3\%) games, which is a bit higher than that in Table \ref{tbl:res1}.

\begin{table*}[htbp]
\caption{Mean and Std of Evaluation Scores after 200M Frames}
\centering
\begin{tabular}{ |c||c c|c|| c c |c|} 
 \hline
    \textbf{Game} & \textbf{Boot-DQN*} &\textbf{Std} &\textbf{Normalized} &\textbf{Boot-DQN+NP} &\textbf{Std} &\textbf{Normalized}\\
 \hline
Alien & 2358.0 & 511.48 & 0.31 & \textbf{3414.0} & 158.19 & 0.46 \\
\hline
Amidar & 401.0 & 24.79 & 0.23 & \textbf{545.6} & 42.77 & 0.31 \\
\hline
Assault & 1207.6 & 270.02 & 1.9 & \textbf{1663.4} & 638.94 & 2.77 \\
\hline
Asterix & 19340.0 & 14808.73 & 2.31 & \textbf{32820.0} & 3263.37 & 3.93 \\
\hline
Asteroids & \textbf{1030.0} & 89.44 & 0.01 & 452.0 & 223.11 & -0.01 \\
\hline
Atlantis & \textbf{873180.0} & 4752.43 & 53.18 & 859960.0 & 8920.0 & 52.36 \\
\hline
Bank Heist & \textbf{566.0} & 108.0 & 0.75 & 0.0 & 0.0 & -0.02 \\
\hline
Battle Zone & \textbf{31600.0} & 3666.06 & 0.84 & 19800.0 & 2227.11 & 0.5 \\
\hline
Beam Rider & 11816.4 & 3616.03 & 0.69 & \textbf{13762.0} & 8137.31 & 0.81 \\
\hline
Bowling & \textbf{30.0} & 0.0 & 0.05 & 9.0 & 9.3 & -0.1 \\
\hline
Boxing & 96.0 & 3.74 & 7.99 & \textbf{98.4} & 1.96 & 8.19 \\
\hline
Breakout & \textbf{403.0} & 7.97 & 13.93 & 398.6 & 7.61 & 13.78 \\
\hline
Centipede & \textbf{2115.0} & 565.45 & 0.0 & 2094.0 & 998.77 & 0.0 \\
\hline
Chopper Command & 880.0 & 97.98 & 0.01 & \textbf{1080.0} & 146.97 & 0.04 \\
\hline
Crazy Climber & 118340.0 & 4412.53 & 4.29 & \textbf{146180.0} & 7049.65 & 5.41 \\
\hline
Demon Attack & 6455.0 & 3324.02 & 3.47 & \textbf{10748.0} & 12470.36 & 5.83 \\
\hline
Double Dunk & -5.6 & 4.8 & 5.91 & \textbf{-1.6} & 1.5 & 7.73 \\
\hline
Enduro & 1171.8 & 164.87 & 1.36 & \textbf{1353.6} & 35.89 & 1.57 \\
\hline
Fishing Derby & -31.4 & 14.88 & 1.14 & \textbf{-26.6} & 7.94 & 1.23 \\
\hline
Freeway & 21.6 & 1.2 & 0.73 & \textbf{33.2} & 0.4 & 1.12 \\
\hline
Frostbite & \textbf{2922.0} & 693.26 & 0.67 & 744.0 & 78.38 & 0.16 \\
\hline
Gopher & 3880.0 & 1048.2 & 1.68 & \textbf{44208.0} & 32534.94 & 20.4 \\
\hline
Gravitar & 70.0 & 140.0 & -0.03 & \textbf{350.0} & 303.32 & 0.06 \\
\hline
Hero & 3010.0 & 0.0 & 0.07 & \textbf{7668.0} & 110.12 & 0.22 \\
\hline
Ice Hockey & -15.0 & 2.45 & -0.31 & \textbf{-4.6} & 1.85 & 0.55 \\
\hline
Jamesbond & \textbf{720.0} & 156.84 & 2.52 & 470.0 & 40.0 & 1.61 \\
\hline
Kangaroo & \textbf{11620.0} & 943.19 & 3.88 & 10920.0 & 2584.88 & 3.64 \\
\hline
Krull & 7088.8 & 1695.02 & 5.14 & \textbf{9295.2} & 461.11 & 7.21 \\
\hline
Kung Fu Master & 2800.0 & 1395.71 & 0.11 & \textbf{23640.0} & 2081.92 & 1.04 \\
\hline
Montezuma Revenge & \textbf{0.0} & 0.0 & 0.0 & \textbf{0.0} & 0.0 & 0.0 \\
\hline
Ms Pacman & 1786.0 & 232.0 & 0.22 & \textbf{3682.0} & 921.88 & 0.51 \\
\hline
Name This Game & \textbf{10072.0} & 1985.32 & 1.35 & 9516.0 & 2868.68 & 1.25 \\
\hline
Pong & \textbf{21.0} & 0.0 & 1.18 & 20.6 & 0.8 & 1.17 \\
\hline
Private Eye & 0.0 & 0.0 & -0.0 & \textbf{100.0} & 0.0 & 0.0 \\
\hline
Qbert & \textbf{22215.0} & 1898.0 & 1.66 & 21475.0 & 2907.71 & 1.6 \\
\hline
Riverraid & 10288.0 & 2686.7 & 0.57 & \textbf{12902.0} & 1582.46 & 0.73 \\
\hline
Road Runner & 43120.0 & 13120.27 & 5.5 & \textbf{60240.0} & 7593.05 & 7.69 \\
\hline
Robotank & 21.6 & 7.39 & 2.0 & \textbf{31.0} & 8.65 & 2.97 \\
\hline
Seaquest &\textbf{17124.0} & 1262.53 & 0.41 & 5852.0 & 1946.34 & 0.14 \\
\hline
Space Invaders & 704.0 & 409.5 & 0.37 & \textbf{844.0} & 537.52 & 0.46 \\
\hline
Star Gunner & 33560.0 & 6499.42 & 3.43 & \textbf{36640.0} & 7793.23 & 3.75 \\
\hline
Tennis & -14.6 & 7.31 & 0.59 & \textbf{-1.2} & 0.4 & 1.46 \\
\hline
Time Pilot & \textbf{5200.0} & 1255.39 & 0.98 & 2860.0 & 1763.63 & -0.43 \\
\hline
Tutankham & 0.0 & 0.0 & -0.07 & \textbf{261.6} & 61.73 & 1.6 \\
\hline
Up N Down & 4432.0 & 153.41 & 0.35 & \textbf{4736.0} & 161.44 & 0.38 \\
\hline
Venture & \textbf{420.0} & 146.97 & 0.35 & 0.0 & 0.0 & 0.0 \\
\hline
Video Pinball & 223341.4 & 104619.78 & 12.64 & \textbf{298420.6} & 61065.4 & 16.89 \\
\hline
Wizard Of Wor & 460.0 & 80.0 & -0.02 & \textbf{1560.0} & 80.0 & 0.24 \\
\hline
Zaxxon & 6140.0 & 1934.53 & 0.67 & \textbf{8880.0} & 1203.99 & 0.97 \\
\hline
\end{tabular}
\label{tbl:appen1}
\end{table*}

\bibliographystyle{abbrv}
\bibliography{main}

\vfill

\end{document}